\documentclass[11pt]{article}

\usepackage[final]{acl}

\usepackage{times}
\usepackage{latexsym}

\usepackage[T2A,T1]{fontenc}

\usepackage[utf8]{inputenc}
\usepackage[russian,english]{babel}

\usepackage{microtype}

\usepackage{inconsolata}

\usepackage{graphicx}

\usepackage{multirow, hhline, booktabs}
\usepackage[utf8]{inputenc}
\usepackage{CJKutf8}
\usepackage{todonotes}
\newcommand{\Note}[2]{} 
\newcommand{\SideNote}[2]{}
\renewcommand{\Note}[2]{\todo[color=#1,size=\small, inline=true]{#2}} 
\renewcommand{\SideNote}[2]{\todo[color=#1,size=\small]{#2}} %
\newcommand{\lingyu}[1]{}
\newenvironment{itemizesquish}{\begin{list}{\labelitemi}{\setlength{\itemsep}{-0.2em}\setlength{\labelwidth}{0.5em}\setlength{\leftmargin}{\labelwidth}\addtolength{\leftmargin}{\labelsep}}}{\end{list}}

\usepackage{xspace}
\newcommand{\dataset}{\textsc{WhoSaidIt}\xspace}
\newcommand{\male}{\textit{male}\xspace}
\newcommand{\female}{\textit{female}\xspace}
\newcommand{\adult}{\textit{adult}\xspace}
\newcommand{\child}{\textit{child}\xspace}
\newcommand{\grandparent}{\textit{elderly}\xspace}
\newcommand{\parent}{\textit{parent}\xspace}
\newcommand{\vegetarian}{\textit{vegetarian}\xspace}
\newcommand{\eatmeat}{\textit{meat-eater}\xspace}
\newcommand{\serious}{\textit{serious}\xspace}

\title{WhoSaidIt: Human-LLM Collaborative Annotation for Text-Based Multilingual Speaker-Attribute Classification}

\author{Lingyu Gao, Will Monroe, David Smith, Meghan Jemison, Jackie Lee\\
Duolingo\\
\texttt{\{lingyu,monroe,david.smith,meghan.jemison,jackie.lee\}@duolingo.com}
}

\begin{document}
\maketitle
\begin{abstract}
Annotating speaker attributes from text is inherently ambiguous, particularly in multilingual settings where demographic and social cues are implicit and culturally variable.
We propose a human-large language model (LLM) collaborative re-annotation framework for stabilizing multilingual speaker-attribute labels under practical resource constraints. Starting from a noisy corpus, we use LLMs to surface recurring annotation rationales through iterative interaction with experts, and apply disagreement-focused sampling for targeted re-annotation.  
Using this framework, we construct \dataset, a multilingual dataset covering nine speaker-attribute labels. We quantify divergence between original and revised annotations, benchmark recent LLMs, and analyze the effect of explicit rationales on model behavior. Our results reveal substantial cross-lingual differences in annotation decisions and demonstrate both the strengths and limitations of LLMs in speaker-attribute classification.

\end{abstract}

\section{Introduction}

Speaker-attribute classification has long been studied in speech processing \cite{10.1016/j.csl.2012.01.008, s21175892, DBLP:conf/odyssey/0007SS024}, where acoustic and prosodic signals are leveraged for speaker assignment and diarization. However, some industry systems operate in text-only environments, where attributes cannot be inferred from voice and must be interpreted from linguistic cues in the text.
These cues may be expressed explicitly or implicitly through morphology, lexical choice, pragmatic framing, and cultural references \citep{https://doi.org/10.1111/josl.12080, DBLP:journals/access/GuimaraesRGRB17}.
While prior work has explored demographic and personality profiling from text, most studies focus on specific languages, longer user-level documents, or supervised modeling settings. Less attention has been paid to how such attributes can be \textbf{consistently defined and annotated across languages} when cues are implicit and culturally variable.

\begin{figure}[t!]
  \centering
  \includegraphics[page=1,width=.98\linewidth]{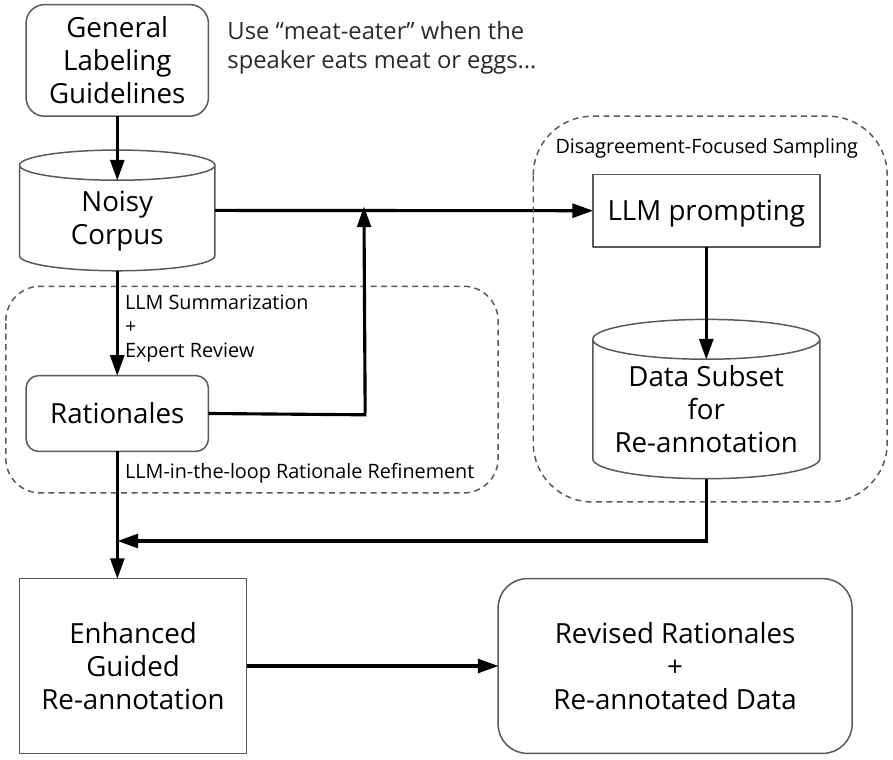}
  \caption{\label{diagram}Diagram for dataset construction pipeline.}
\end{figure}

Unlike acoustic features, linguistic cues are often implicit and governed by language-specific conventions. The same concept may be expressed differently across languages, and its interpretation may depend on local legal or social norms.
For example, references to driving may be interpreted as evidence of ``adulthood'' in some countries, while in others teenagers are legally permitted to drive.
Such cross-cultural asymmetries can create ambiguous decision boundaries and substantial disagreement among annotators \cite{10.1145/3593013.3594098, lee-etal-2024-exploring-cross}. 
Consequently, constructing multilingual datasets for speaker-attribute classification requires iteratively defining and stabilizing subjective sociolinguistic categories across languages, as reliable label definitions often emerge only through empirical examination of diverse instances.

To address these challenges, we propose a human-LLM collaborative re-annotation framework designed for subjective, multilingual labeling under practical resource constraints. 
As shown in Figure~\ref{diagram}, starting from a large noisy corpus, we employ LLMs to summarize cross-lingual annotation rationales. These rationales are manually consolidated by expert annotators into unified guidelines and used to guide targeted re-annotation. To efficiently allocate annotation effort, we apply disagreement-focused sampling to prioritize high-information and ambiguous instances. The refined rationales are further incorporated into LLM prompts to support interactive quality control, enabling iterative feedback between models and human experts.

We instantiate this framework in the construction of \dataset, a multilingual dataset for text-only speaker-attribute classification covering eleven languages and nine binary attributes.\footnote{We release a smaller public subset covering five languages: English, Spanish, Italian, Korean, and Chinese. The released portion contains 3,600 examples in total across 9 labels and is available at \url{https://github.com/duolingo/whosaidit}.}

Each instance is annotated for speaker characteristics including gender, age group, parental status, dietary preference, and personality-related traits. Table~\ref{example_annotation} presents illustrative examples. We quantify the divergence between original and revised annotations, benchmark several recent LLMs on the re-annotated data, and further analyze how explicit rationales influence model behavior. Our results show that while LLMs can assist in identifying overlooked cases and clarifying decision boundaries, they remain sensitive to surface cues and struggle with subtle pragmatic distinctions. Collectively, these findings provide practical guidance for designing LLM-in-the-loop annotation workflows in multilingual industrial settings.

\begin{CJK}{UTF8}{min}
\begin{table}[!t]
\small
\centering
\begin{tabular}{p{0.65\linewidth}p{0.22\linewidth}}\toprule
Sample Text & Label(s) \\\midrule
We are boys. & \male, \child \\
私の孫たちはたくさん遊んでいます。(My grandchildren are playing a lot.) & \grandparent, \adult \\
\foreignlanguage{russian}{Моя жена умная.} (My wife is smart.) & \adult \\
Je suis végétarienne, comme ma mère. (I’m a vegetarian, like my mother.) & \female, \vegetarian \\
\bottomrule
\end{tabular}
\caption{\label{example_annotation} Examples from \dataset.}
\end{table}
\end{CJK}

\section{Related Work}
\paragraph{Subjective Multilingual Annotation and Human-LLM Collaboration.}
Subjective NLP tasks often produce systematic annotator disagreement rather than random noise, especially in multilingual and cross-cultural settings where decision boundaries depend on local norms, pragmatic conventions, and implicit linguistic cues \citep{joshi-etal-2016-cultural, plank-2022-problem, sandri-etal-2023-dont, Cabitza_Campagner_Basile_2023, 10.1145/3593013.3594098, lee-etal-2024-exploring-cross}. This challenge is central to our setting: speaker attributes are often weakly signaled in text, and the same cue may support different inferences across languages. We therefore treat disagreement as evidence that label definitions and operational rationales must be clarified.

Recent work uses LLMs to support annotation through active learning, verification-based correction, and interactive human-LLM interfaces \citep{DBLP:conf/pkdd/KholodnaJKGG24, 10.1145/3613904.3641960, kim-etal-2024-meganno}. These approaches typically use LLMs to accelerate or improve labeling under a largely fixed schema. Complementary work studies LLMs and annotation guidelines directly: \citet{bibal-etal-2025-automating} use LLMs to improve entity-recognition guidelines, \citet{fonseca-cohen-2024-large} study whether LLMs can follow concept annotation guidelines, and recent work simulates pilot annotation with LLMs to refine instructions \citep{kim2026diziner}. Our framework differs by using LLMs as analytical tools rather than replacement annotators or autonomous guideline writers: model outputs surface recurring cross-lingual rationales, which human experts then consolidate into revised guidelines.

Our design also addresses concerns that LLM suggestions can shape human judgments and downstream label distributions \citep{choi-etal-2024-llm, schroeder-etal-2025-just}. Primary annotators therefore make independent judgments before model outputs are considered. We use model-annotation disagreement mainly for targeted sampling and expert quality control, reducing anchoring risks while still benefiting from LLMs' ability to reveal overlooked patterns.

\paragraph{Speaker-Attribute Classification from Text.}
In NLP, related work is usually framed as author profiling, authorship attribution, or demographic and personality prediction from text, often using long user histories, metadata, or platform-specific stylistic features \citep{https://doi.org/10.1111/josl.12080, verhoeven-etal-2016-twisty, DBLP:journals/access/GuimaraesRGRB17, sari-etal-2018-topic, 10.1016/j.eswa.2022.117140}. Shared tasks such as PAN provide benchmarks for multilingual gender and age prediction \citep{DBLP:conf/clef/PardoCRPSD15}, but rely on user-level aggregation rather than sentence-level inference.
Our setting,
motivated by a production content-labeling workflow (Section~\ref{sec:deployment}), differs in that attributes must be inferred from a single short utterance, without user history, metadata, images, or acoustic signals. We cover nine attributes across eleven typologically diverse languages, emphasizing cross-lingual consistency rather than monolingual style modeling. Because the evidence is sparse, implicit, and culturally contingent, dataset construction requires stabilizing annotation guidelines; our contribution is therefore both a benchmark and a human-LLM collaborative re-annotation framework for refining rationales and selectively revising noisy labels.

\section{Speaker-attribute Text Classification}\label{sec:task}

Given an input text $x$, our objective is to predict a sequence of binary labels 
$y = (y_1, y_2, \ldots, y_K)$, where each label $y_k \in \{0,1\}$ indicates the presence of a particular speaker attribute $a_k$. 

A single input may carry multiple positive labels simultaneously, and the general label definitions are shown in Table~\ref{label_def}.\footnote{See Appendix~\ref{app:annot_rationale} for illustrative detailed annotation rationales.} The label set covers gender (\male, \female), age group (\child, \adult, \grandparent), parental status (\parent), dietary preference (\eatmeat, \vegetarian), and personality trait (\serious). 

Most textual inputs can be uttered by any speaker. While each label is formulated as an independent binary classification problem rather than a single multi-label setting, certain label dependencies exist. For example, \parent and \grandparent will co-occur with \adult. However, mutually exclusive pairs such as \male and \female, \eatmeat and \vegetarian, and \parent and \grandparent (as defined in this task) should not be given a label of 1 simultaneously.

\begin{table}[!t]
\small
\centering
\begin{tabular}{p{0.17\linewidth}p{0.72\linewidth}}\toprule
Label & Definition \\\midrule
\male & The speaker self-identifies as male. \\
\female & The speaker self-identifies as female. \\\midrule
\child & The speaker is a child or teenager. \\
\adult & The speaker is an adult, or the sentence involves ``adult'' themes (e.g., alcoholic or caffeinated beverage consumption). \\
\grandparent & The speaker is elderly or identified as a grandparent. \\\midrule
\parent & The speaker states or is implied to have children, but does not indicate that the speaker is elderly or a grandparent. \\\midrule
\eatmeat & The speaker eats meat, poultry, seafood, or eggs (dairy products are excluded).  \\
\vegetarian & The sentence specifically states (or implies) the speaker is vegetarian. \\\midrule
\serious & The sentence deals with serious or negative themes, such as death, illness, crime, or sadness. \\\bottomrule
\end{tabular}
\caption{\label{label_def} General definitions of labels used in \dataset.}
\end{table}

\section{\dataset: Dataset Construction}

We construct \dataset, a multilingual dataset spanning 11 languages: Japanese, Portuguese, English, Spanish, German, French, Italian, Korean, Russian, Turkish, and Chinese. The task covers nine speaker-attribute labels.

Starting from a noisy corpus, we use LLMs to expand our guidelines (rationales) and identify a smaller, informative subset for re-annotation, since full relabeling is impractical under operational resource constraints.

\subsection{Initial Noisy Dataset}

Our starting point is a large multilingual corpus of approximately 195,000 short textual inputs, spanning 22 languages including the 11 languages above. Each language set was annotated over an extended period of time by native or proficient language experts.
However, the resulting annotations exhibit considerable noise, characterized by the following features:

\begin{itemizesquish}
    \item \textbf{Partial coverage:} Most data was annotated for only a subset of the nine target labels.
    \item \textbf{Label imbalance:} The dataset is highly skewed, with few positive instances for each attribute.
    \item \textbf{Broad guidelines:} The initial annotation instructions contained general rationales for each label but remained under-specified (as stated in Table~\ref{label_def}), requiring annotators to rely heavily on personal judgment.
    \item \textbf{Annotation noise:} Inconsistent interpretations were observed across languages, and occasional human errors remained even  with a verifying procedure in our annotation process.
\end{itemizesquish}

While this large-scale corpus provided wide linguistic coverage, label variability and ambiguity motivated the development of a second, LLM-assisted re-annotation process described below.

\subsection{LLM-in-the-loop Rationale Refinement}
\label{sec:original_sample}

The broad guidelines for the initial annotations reflected the intrinsic difficulty of the task: it is nearly impossible to anticipate all possible linguistic realizations of speaker attributes before examining real data. To address this, we applied an LLM (GPT-4o-08-06; \citealp{DBLP:journals/corr/abs-2410-21276}) to help summarize the annotation guidelines.

For each language, we randomly sampled 
up to 50 positive and 50 negative examples 
from the initial corpus and iteratively interacted with the LLM in English to analyze the sampled sentences and surface recurring linguistic cues and decision patterns associated with each attribute. The LLM outputs were treated as analytical suggestions rather than authoritative decisions. Language experts manually reviewed and consolidated these observations into a unified cross-lingual guideline document.

For example, for label \adult, the initial guideline focused on explicitly adult-associated content (e.g., alcohol, caffeinated beverages, references to a spouse, or other content inappropriate for children). After iterative refinement, 
we expanded the adult definition to include professional, financial, and civic responsibilities while adding negative constraints to prevent overgeneralization.

These refined guidelines promote cross-lingual consistency while accounting for cultural variation (e.g., differing legal drinking ages). When disagreements arise, they reflect an averaged annotation pattern across languages, which may revise original labels. For example, neutral descriptions such as ``This is a vegetarian restaurant.'' are no longer labeled \vegetarian.

\subsection{Disagreement-Focused Sampling} \label{sec:sampling}

The initial corpus contains annotation noise, and the refined guidelines in Section~\ref{sec:original_sample} further clarify or revise the original decision boundaries, potentially creating inconsistencies with existing labels. Because re-annotating the entire corpus is impractical given its size and imbalance, we adopt a targeted strategy to identify high-information instances for correction.

From the initial corpus, we construct label-wise roughly balanced development and test splits, referred to as the \textit{intermediate dataset}. The exploratory sample in Section~\ref{sec:original_sample} was used only for rationale refinement. In contrast, the intermediate dataset constructed here is used to analyze disagreement between LLM predictions and the original annotations. Using prompts derived from the refined rationales, we obtain LLM predictions on this split and categorize each instance as true positive, true negative, false positive, or false negative relative to the original annotations.\footnote{Detailed statistics and prediction results for the intermediate dataset are provided in Appendix~\ref{app:inter_data}. A prompt example for \eatmeat at this stage is shown in Table~\ref{tab:prompt_eatmeat_initial}.}

We then construct a \textit{disagreement-sampled subset} by oversampling model-annotation disagreement cases (false positives and false negatives), treating them as heuristic indicators of ambiguous cases, where discrepancies may reflect annotation errors or shifts in guideline interpretation. We sample approximately twice as many disagreement cases as agreement cases to prioritize high-information examples for re-annotation. Due to resource constraints, we restrict this disagreement-sampled subset to 11 languages.\footnote{The exact distribution may vary slightly by language depending on available data in each category.}

\subsection{Enhanced Guided Re-annotation}
\label{sec:reannotation}

The disagreement-focused subset was re-annotated using the refined rationales as primary guidelines. For each language, a single trained annotator labeled all instances in the subset. To manage label complexity, the re-annotated data was divided into two batches: one including \male, \female, \child, \adult, \grandparent, and \parent, and the other including \eatmeat, \vegetarian, and \serious.

During annotation, annotators were encouraged to flag ambiguous or borderline cases. Such cases were discussed within the annotation team, and recurring ambiguities led to clarifications in the shared rationale document to standardize interpretation. Revisions at this stage primarily clarified objective descriptions, negative statements, time-dependent expressions, and the treatment of questions and sentence fragments.

Taking diet preference as an example, we clarified several corner cases, including past-tense statements (e.g., “I was vegetarian 10 years ago”), third-person descriptions and questions, and negative statements that still imply meat consumption. These refinements reduced ambiguity and standardized cross-lingual interpretation.

To mitigate the limitations of single-primary-annotator labeling, a senior expert performed quality control through random audits and targeted review of selected LLM-annotator disagreement cases. When discrepancies were identified, the senior expert made the final determination according to the refined guidelines. In cases where ambiguity persisted, we followed a conservative adjudication principle: unless clear evidence warranted revision, the annotator’s original label was retained. As a result, every instance in the dataset received a single finalized binary label (0/1) per attribute. Boundary cases were retained rather than filtered. We provide additional details on this resource-aware quality-control protocol in Appendix~\ref{app:annotation_qc}.

\section{Data Analysis}

\subsection{Data Statistics}

\begin{table}[!t]
\small
\centering
\begin{tabular}{lrrrr}\toprule
\multicolumn{2}{l}{\# data} & True & False & Total \\\midrule
\multirow{2}{*}{gender} & \male & 188 & 2,812 & 3,000 \\
 & \female & 124 & 2,876 & 3,000 \\\midrule
\multirow{3}{*}{age group} & \child & 216 & 2,784 & 3,000 \\
 & \adult & 996 & 2,004 & 3,000 \\
 & \grandparent & 113 & 2,887 & 3,000 \\\midrule
parental status  & \parent & 141 & 2,859 & 3,000 \\\midrule
\multirow{2}{*}{diet preference} & \eatmeat & 326 & 2,036 & 2,362 \\
 & \vegetarian & 158 & 2,204 & 2,362 \\\midrule
personality traits & \serious & 463 & 1,899 & 2,362\\\bottomrule
\end{tabular}
\caption{\label{statistics} Data statistics of the re-annotated data.}
\end{table}

\begin{table*}[!t]
\small
\centering
\begin{tabular}{p{0.15\linewidth}|p{0.06\linewidth}p{0.06\linewidth}|p{0.06\linewidth}p{0.06\linewidth}p{0.06\linewidth}|p{0.06\linewidth}|p{0.09\linewidth}p{0.06\linewidth}|p{0.06\linewidth}}\toprule
\multirow{2}{*}{} & \multicolumn{2}{c|}{gender} & \multicolumn{3}{c|}{age group} &   & \multicolumn{2}{c|}{diet} &  \\
 & \male & \multicolumn{1}{l|}{\female} & \child & \multicolumn{1}{l}{\adult} & \multicolumn{1}{l|}{\grandparent} & \parent & \eatmeat & \multicolumn{1}{l|}{\vegetarian} & \serious \\\midrule
DeepSeek V3 & 74.7 & 67.0 & 38.5 & 79.4 & 70.2 & 77.7 & 82.5 & 58.6 & 61.7 \\
Gemini 2.5 Flash & 83.2 & 77.9 & 40.4 & 77.5 & 64.0 & 70.1 & 83.0 & \textbf{86.6} & 57.9 \\
GPT 4.1 & \textbf{85.0} & 88.9 & 42.5 & 83.0 & 83.5 & \textbf{86.6} & \textbf{85.7} & 72.9 & 63.2 \\
Claude 3.7 Sonnet & 84.8 & \textbf{93.3} & \textbf{43.9} & \textbf{83.3} & \textbf{89.4} & 85.5 & 83.4 & 83.0 & \textbf{66.1}\\
\bottomrule
\end{tabular}
\caption{\label{results} F1 (\%) for the positive class on the re-annotated test set.}
\end{table*}

\begin{table*}[!t]
\small
\centering
\begin{tabular}{p{0.15\linewidth}|p{0.06\linewidth}p{0.06\linewidth}|p{0.06\linewidth}p{0.06\linewidth}p{0.06\linewidth}|p{0.06\linewidth}|p{0.09\linewidth}p{0.06\linewidth}|p{0.06\linewidth}}\toprule
\multirow{2}{*}{} & \multicolumn{2}{c|}{gender} & \multicolumn{3}{c|}{age group} &   & \multicolumn{2}{c|}{diet} &  \\
 & \male & \multicolumn{1}{l|}{\female} & \child & \multicolumn{1}{l}{\adult} & \multicolumn{1}{l|}{\grandparent} & \parent & \eatmeat & \multicolumn{1}{l|}{\vegetarian} & \serious \\\midrule
 DeepSeek V3 & 55.7 & 50.4 & 26.2 & 74.4 & 50.0 & \textbf{83.1} & 63.4 & 70.0 & 63.6 \\
Gemini 2.5 Flash & 53.1 & 56.4 & 41.1 & \textbf{79.2} & 37.0 & 59.5 & 24.0 & 72.4 & 62.2 \\
GPT 4.1 & 73.2 & 71.1 & 38.7 & 69.1 & 64.9 & 69.1 & 51.7 & 68.8 & \textbf{64.7} \\
Claude 3.7 Sonnet & \textbf{78.6} & \textbf{87.2} & \textbf{45.0} & 59.2 & \textbf{83.2} & 82.8 & \textbf{71.5} & \textbf{76.9} & 28.1$^\dagger$ \\\bottomrule
\end{tabular}
\caption{\label{results_delta_rm_rationales} F1 (\%) for the positive class on the re-annotated test set after removing rationales from prompts. $\dagger$: not directly comparable; see Section~\ref{sec:rationale}.}
\end{table*}

Table~\ref{statistics} reports the size of the final re-annotated data (dev and test combined), corresponding to the disagreement-focused subset after expert review (Section~\ref{sec:reannotation}). The instances are unchanged; only their labels were revised.

Because sampling was performed separately from the intermediate development and test splits (Section~\ref{sec:sampling}), the re-annotated dataset preserves the same split structure. Data sizes differ across label groups due to batching and resource constraints, with approximately 3,000 instances per demographic attribute and 2,362 per diet and personality attribute.

\subsection{Comparison with Original Labels}

To quantify how much the refined guidelines alter labeling outcomes, we compute Cohen’s $\kappa$ coefficient for each attribute between the original corpus labels and the final re-annotated labels for the same instances in the disagreement-focused subset. This analysis measures the extent of label revision introduced by the guideline refinement.

We observe that agreement varies widely across both attributes and languages.\footnote{Detailed results are provided in Table~\ref{kappa} in the Appendix.} Higher $\kappa$ values (e.g., for \grandparent and \eatmeat) indicate that the new guidelines largely preserve the original labeling decisions, whereas lower values (e.g., for \child and \serious) reflect substantial reinterpretation under the refined guidelines. Notably, because of differences in how annotators infer speaker gender, the $\kappa$ values for \male and \female are high in Japanese and Italian (\textgreater \ 0.9) but very low in English and Chinese (\textless \ 0.1).

\section{Experimental Setup}

This benchmarking stage is distinct from the disagreement-focused sampling procedure in Section~\ref{sec:sampling}. Here, LLMs are evaluated as classifiers on the finalized re-annotated dataset, rather than used as heuristic tools for sampling.

All reported precision, recall, and F1 scores are computed on the held-out test set using the finalized re-annotated labels from Section~\ref{sec:reannotation}. The dev set is used solely for prompt refinement (e.g., wording adjustments and boundary inspection). In-context examples are manually written rather than copied from the dataset; when development instances inform prompt design, they are rewritten to avoid direct reuse.

For certain labels, such as \eatmeat, the prompt itself is structured as a step-by-step decision procedure: first determining whether a food item appears in the given text, then checking whether that food contains meat, before finally making a judgment.

We experiment with DeepSeek V3 \cite{DBLP:journals/corr/abs-2412-19437}, Gemini 2.5 Flash \cite{DBLP:journals/corr/abs-2507-06261}, GPT-4.1 (2025-04-14) \cite{openai2025gpt41}, and Claude 3.7 Sonnet \cite{Claude3S}. We set temperature to 0 for more deterministic and reproducible outputs as baseline results.\footnote{Some closed-source models may still produce slight output variation under this setting.} 

\section{Results and Analysis}\label{sec: results}
\subsection{Results}

Results are shown in Table~\ref{results}.\footnote{Because labels are highly imbalanced and the positive class corresponds to the attribute of interest, we report positive-class precision, recall, and F1 as our primary metrics.} Claude 3.7 Sonnet achieves the best overall performance, leading on 5 of 9 labels. GPT-4.1 ranks second with comparable results at lower cost.

In general, the LLMs perform well except on \child and \serious, possibly because these labels involve greater subjectivity and pragmatic nuance, making them harder to classify consistently. Because the subset emphasizes disagreement cases, it also represents a challenging evaluation setting.

When evaluated on the full intermediate test set, using re-annotated labels for the sampled subset and original labels for the remaining instances, performance increases substantially (e.g., F1 exceeding 0.7 for \child and 0.8 for \adult), mainly due to a sharp increase in true positives.

\subsection{Impact of Rationales}\label{sec:rationale}

We perform an ablation study by removing detailed annotation rationales from the evaluation prompts (i.e., expanded operational rules and corner cases), retaining only the general label definitions in Table~\ref{label_def} and a single in-context example for format control. Results are shown in Table~\ref{results_delta_rm_rationales}.

Removing rationales substantially changes model behavior. Claude 3.7 Sonnet remains the strongest model on 6 of the 8 directly comparable labels, while Gemini 2.5 Flash experiences larger performance drops on 5 of the 9 labels when rationales are removed, suggesting weaker alignment with the label-specific decision rules encoded in our rationales. For \serious, the three directly comparable models obtain slightly higher F1 after rationales are removed, indicating that rationales can sometimes introduce additional ambiguity or overconstraint.

The Claude \serious score is treated separately because the no-rationale run exhibited a prompt-formatting artifact: in 76\% of cases, Claude's rationale referred to the in-context demonstration rather than the target text, a behavior not observed for the other three models on the same prompt. The reported value$^\dagger$ excludes these outputs and is not directly comparable to the full-test-set scores. Even on this subset, Claude remains weak on \serious, often missing non-emotive cues such as political, medical, legal, safety, or protest-related content.

\section{Real-world Deployment}\label{sec:deployment}

Although our primary contribution is methodological, we operationalized this framework in an industrial content-labeling workflow that assigns speaker-attribute tags to short texts for consistency between the characters in the Duolingo app and their utterances in the target learning language.

Before adoption of the framework described here, these tags were manually annotated; a labeling workload that previously required approximately two weeks of annotator effort can now be completed in approximately three hours using classification prompts derived from our refined rationales. 
Approved for production use in July 2025, this rationale-derived prompting workflow has been in continuous use since.

This operational use illustrates a property of the framework that is difficult to evaluate through benchmarks alone: the same human-readable rationale serves both as an annotation guideline for future annotation and quality control, and as a prompt specification for LLM-based classification. When an annotation boundary shifts, domain experts revise the rationale once, and the change propagates to both human guidance and model prompts. This is especially useful in multilingual settings: although our re-annotation and evaluation focus on 11 languages and 9 labels, the workflow originated from a 22-language corpus and is expected to expand with product coverage. By using rationales as the shared interface between human annotation and LLM classification, the workflow can adapt to changing guidelines and new language-label combinations without requiring high-quality training data for each combination, a burden exacerbated by highly imbalanced labels in real-world data distributions.

\section{Discussion: Human-LLM Collaboration in Annotation}\label{sec:discussion}

Human annotation is costly and inconsistent, particularly in multilingual settings. While LLMs improve efficiency, they may introduce anchoring bias \citep{choi-etal-2024-llm, schroeder-etal-2025-just}; we therefore reveal model outputs only after annotators make their initial judgments, preserving independent human decisions while enabling structured comparison.

We analyze human-LLM interaction in two practical annotation settings. In the first, annotators complete large batches of multi-label annotations (over 10,000 instances), reflecting routine annotation workflows. In the second setting, during re-annotation, an expert reviews the disagreement-focused subset, adjudicates labels according to the refined guidelines, and analyzes recurring model error patterns. This analysis provides qualitative feedback on model behavior but does not alter the annotation guidelines or benchmark prompts. The resulting adjudicated labels serve as the reference for subsequent analysis.

\paragraph{LLMs can help correct human omissions.}
Using the finalized re-annotated labels as reference, we compare the original large-batch corpus annotations (Setting 1) to the updated labels on the disagreement-focused subset. The original routine annotations exhibit high precision (above 0.9 for 8 out of 9 labels) but substantially lower recall, often around 0.5 and exceeding 0.8 only for \male and \female. In several instances, the LLM correctly identified examples that human annotators initially overlooked, e.g., ``\textit{Questo non sembra vegetariano!}'' (This doesn’t look vegetarian!) for \vegetarian, suggesting its utility for recall-oriented quality checks.

\paragraph{LLMs still struggle with context and inference.} Annotator feedback reveals two recurring issues: (i) lexical overreliance, where models overgeneralize from surface cues (e.g., assigning \female based on gendered nouns or labeling ``mom'' as \parent in third-person contexts), and (ii) context hallucination, where unsupported relations or intent are inferred. Although LLM-generated rationales clarify decision boundaries, they reveal limited pragmatic sensitivity. Subtle cues of tone, intent, and social context remain difficult to model, underscoring the need for human verification in nuanced tasks.

\section{Conclusion}

We presented a human-LLM collaborative framework for multilingual speaker-attribute classification, using LLMs to identify cross-lingual annotation patterns, refine rationales, and prioritize high-information disagreement cases for re-annotation. With this framework, we built \dataset, a nine-label multilingual corpus based solely on textual cues, and benchmarked four recent LLMs on the refined data. The resulting prompts have also been deployed in a production content-labeling workflow, reducing a previously manual labeling process from approximately two weeks of annotator effort to about three hours. Results show that LLMs can effectively complement human annotation by detecting overlooked signals, yet still struggle with subtle pragmatic and contextual inference. A data subset is released to support future research, with a broader release possible where data policy permits. Future work could investigate finetuned and adapter-based models for language-label combinations with sufficient data, as well as automatic prompt optimization approaches that adapt prompts to language- and label-specific annotation challenges.

\section*{Limitations}

While our study highlights the potential of LLM-assisted annotation, several limitations remain. 

First, using an LLM-in-the-loop setup may introduce bias toward the model's own decision boundaries, potentially aligning the dataset with specific model behaviors. We mitigate this by using model predictions only as a sampling heuristic, keeping primary re-annotation independent of model outputs, relying on human adjudication for final labels, and benchmarking multiple LLMs. Nevertheless, the resulting subset should be interpreted as a challenging disagreement-focused benchmark rather than an unbiased sample of all production data.

Second, we acknowledge that some label boundaries remain subjective or context-dependent. Despite careful definition and rationale design, ambiguity in sociolinguistic categories (e.g., \child\ or \serious) persists, which can lead to inconsistency across annotators and models.

Third, due to resource constraints, re-annotation relies on one primary annotator per language, followed by senior-expert quality control through random audits and targeted review. This limits our ability to compute inter-annotator agreement and fully capture cross-linguistic variation. The design reflects available human resources rather than an ideal annotation setup; our framework can in principle accommodate multiple annotators per language when resources permit.

Finally, our evaluation focuses on prompting-based LLM classifiers. We did not finetune models because the current production setting requires maintainable rationales across many imbalanced language-label combinations. Finetuned or adapter-based models may improve accuracy in sufficiently resourced languages and are an important direction for future work.

\section*{Ethical Considerations}\label{sec:ethics}
 
This work involves speaker-attribute inference, including gender, age-related labels, parental status, dietary preference, and personality-related traits. We acknowledge that techniques in this space can in principle be misused for profiling or surveillance of real users. Our task design, however, targets a narrower setting: improving labeling consistency for fictional or stock-image speakers
paired with short textual content. 
Our dataset consists only of short fictional texts that do not correspond to real individuals. While the linguistic cues used to express speaker attributes can be implicit or culturally contextualized, the label definitions are intended to capture task-specific textual evidence rather than latent identity inferences: each label is assigned post hoc based only on cues within the text itself. Systems trained on this dataset could assist with searching large corpora for language that expresses particular speaker attributes, but the resource is not practical for deanonymization, demographic profiling, or inferring other private identifying information.

Although our texts are fictional, we note the label distributions are imbalanced (Table~\ref{statistics}) and LLM performance varies substantially across labels (Table~\ref{results}). Therefore, models trained on these data may still exhibit uneven reliability when applied to real-world text, and should not be used to infer attributes of real individuals without language-specific evaluation and human review of ambiguous cases.
 
Our label schema also simplifies socially complex categories.
For example, gender labels are assigned only when the text provides specific cues supporting a particular speaker gender. In the dataset we release, inferrable gender is limited to \male and \female, but this distribution of textual evidence in the dataset should not be interpreted as an exhaustive or complete representation of gender identity; furthermore, most sentences in our dataset are gender-neutral and do not carry a gender label.

\section*{Acknowledgments}

We thank the Duolingo annotators who contributed to the data annotation process and provided feedback that helped improve the labeling guidelines. We are especially grateful to Erika Puricelli for her annotation work and for providing expert feedback on model behavior in the human-LLM annotation setting. We also thank Elise Kimber and Elisha Sum for their support in coordinating the annotation process, discussing labeling decisions, and refining the annotation guidelines and rationales. We thank Jerry Lan for building the workflow infrastructure that supported efficient prompt iteration. We thank our team leads, Isaac Andersen and Ari Moline, for supporting and encouraging this research alongside our product work. Finally, we thank Andrew Hogue and Klinton Bicknell for their support throughout the release process for the data, paper, and prompts.

\bibliography{custom, anthology}

\clearpage
\appendix
\section{Appendix}\label{sec:appendix}

\subsection{Intermediate Balanced Dev/Test Split Statistics}\label{app:inter_data}

The data statistics for intermediate dev/test split are listed in Table~\ref{inter_statistics}.

Our intermediate classifier results are shown in Table~\ref{baseline}, which illustrates discrepancies between the original annotations and model predictions. It is worth noting that \adult and \child are difficult due to different interpretations. For \male and \female, the gap mainly comes from the decision on whether we assume the speaker's gender based on their partner. The refined rationales clarify that such assumptions should not be made unless explicitly supported by the utterance or context.

\subsection{Details of LLM-in-the-loop Rationale Refinement}\label{app:llm_refinement}

The rationale refinement stage described in Section~\ref{sec:original_sample} was conducted through exploratory, conversational interactions with the LLM. The objective was not to automate labeling, but to surface recurring linguistic cues and potential decision patterns from sampled examples. Interactions were iterative: an expert engaged with the model to generate initial summaries and issue follow-up queries to probe ambiguous or under-specified cases. The resulting observations were subsequently discussed among the annotation team and consolidated into unified cross-lingual guidelines, enabling incremental clarification of ambiguous cases and refinement of guideline distinctions.

All interactions were conducted in English. The original sentences were provided in their respective languages, while the model was instructed to summarize patterns in English to maintain a unified cross-lingual abstraction layer. We did not systematically compare language-specific prompting strategies, as the objective of this stage was guideline consolidation rather than optimizing model-specific cultural representations.

\begin{table}[]
\small
\centering
\begin{tabular}{lrrr}\toprule
\multicolumn{2}{l}{\# data} & dev & test \\\midrule
\multirow{2}{*}{gender} & \male & 2,169 & 2,169 \\
 & \female & 2,025 & 2,026 \\\midrule
\multirow{3}{*}{age group} & \child & 4,684 & 4,684 \\
 & \adult & 6,179 & 6,179 \\
 & \grandparent & 466 & 466 \\\midrule
parental status  & \parent & 1,817 & 1,817 \\\midrule
\multirow{2}{*}{diet preference} & \eatmeat & 3,043 & 3,044 \\
 & \vegetarian & 722 & 722 \\\midrule
personality traits & \serious & 4,047 & 4,048\\\bottomrule
\end{tabular}
\caption{\label{inter_statistics} Data statistics of intermediate data.}
\end{table}

\begin{table}[]
\small
\centering
\begin{tabular}{lrrrr}\toprule
 (\%) & \multicolumn{1}{c}{\textbf{precision}} & \multicolumn{1}{c}{\textbf{recall}} & \multicolumn{1}{c}{\textbf{F1}} & \multicolumn{1}{c}{\textbf{accuracy}} \\\midrule
\male & 94.5 & 55.9 & 70.3 & 84.7 \\
\female & 95.3 & 45.6 & 61.7 & 83.9 \\
\child & 95.4 & 30.2 & 45.9 & 64.4 \\
\adult & 73.2 & 82.1 & 77.4 & 76.0 \\
\grandparent & 99.0 & 88.7 & 93.6 & 97.0 \\
\parent & 99.5 & 88.1 & 93.4 & 96.9 \\
\eatmeat & 92.2 & 97.5 & 94.8 & 96.7 \\
\vegetarian & 99.3 & 80.9 & 89.2 & 95.2 \\
\serious & 88.2 & 92.1 & 90.1 & 91.4 \\\bottomrule
\end{tabular}
\caption{\label{baseline} Baseline results on intermediate test set.}
\end{table}

\subsection{Data Flow and Annotation Stages}

For clarity, we summarize the relationships between the different data stages used in this work.

\paragraph{Initial corpus.}
Our starting point is a multilingual corpus of approximately 195,000 instances with original (noisy) labels.

\paragraph{Exploratory refinement sample (Section~\ref{sec:original_sample}).}
For guideline refinement, we randomly sampled up to 50 positive and 50 negative instances per label per language. These examples were used only for iterative LLM-assisted rationale refinement and were not re-annotated as part of the final benchmark.

\paragraph{Intermediate dataset (Section~\ref{sec:sampling}).}
From the initial corpus, we constructed label-wise roughly balanced development and test splits (Table~\ref{inter_statistics}). This intermediate dataset retained the original corpus labels and was used for model-annotation disagreement analysis.

\paragraph{Disagreement-focused subset.}
A subset of the intermediate dataset was selected by oversampling model-annotation disagreement cases for each attribute. Selection was performed independently per label; thus, an instance could be included due to disagreement on one attribute.

\paragraph{Final re-annotated dataset.}
All instances in the disagreement-focused subset were subsequently re-annotated in batches covering multiple attributes (Section~\ref{sec:reannotation}). Although sampling was triggered by disagreement on specific labels, re-annotation was applied to all attributes within the corresponding batch. Consequently, the final re-annotated dataset contains updated labels for multiple attributes per instance, not only for the attribute that triggered sampling. The instances are identical to those in the sampled subset; only their labels were revised. All statistics reported in Table~\ref{statistics} refer to this re-annotated dataset.

\subsection{Annotation Quality Control under Resource Constraints}\label{app:annotation_qc}

Full multi-annotator labeling with inter-annotator agreement is a standard way to estimate annotation reliability, but it is difficult to scale in multilingual settings where each language requires qualified annotators and positive labels are sparse. Prior work shows that dataset construction often balances reliability against annotation cost through guideline refinement, validation, correction, and adjudication rather than exhaustive double annotation \citep{artstein-poesio-2008-survey,dligach-palmer-2011-reducing,klie-etal-2024-analyzing}. Resource-aware designs, including single primary annotations with validation subsets, moderator review, or task-specific quality checks, have also been used in large-scale and expert-domain datasets \citep{kwiatkowski-etal-2019-natural,ogorman-etal-2021-ms,loukachevitch-etal-2021-nerel}. Our multilingual speaker-attribute setting is similarly constrained: it spans eleven languages and nine labels, and reliable annotation requires both language competence and familiarity with refined attribute guidelines.

We therefore used a targeted quality-control protocol rather than full double annotation. Re-annotation was guided by a versioned cross-lingual rationale, and labels were annotated in batches to reduce decision complexity. Annotators flagged ambiguous or borderline cases, which were discussed by the team; recurring issues led to updates in the shared rationale document. LLM disagreement sampling provided an additional quality signal by identifying model-annotator divergences for targeted review. A senior expert then performed random audits and adjudicated selected disagreement cases according to the refined guidelines. When ambiguity persisted, we followed a conservative principle: unless clear evidence warranted revision, the primary annotator's label was retained.

This protocol is not a substitute for full multi-annotator agreement measurement, which we report as a limitation. However, it concentrates human effort on high-information and ambiguous cases, keeps final decisions grounded in auditable rationales, and propagates guideline updates consistently across languages and batches. As an additional coarse check on label stability, we report Cohen's $\kappa$ between the original noisy labels and the re-annotated labels in Table~\ref{kappa}.

\subsection{Disagreement-based Sampling and Potential Model Bias}

Disagreement-focused sampling used predictions from Claude 3.7 Sonnet to identify instances where model predictions differed from the original corpus annotations. These model-annotation disagreement cases were oversampled to construct the subset selected for human re-annotation. This design intentionally creates a difficult benchmark enriched for ambiguous or potentially mislabeled examples, but it also introduces a possible circularity because Claude 3.7 Sonnet is later evaluated on the same benchmark.

We mitigate this risk in three ways. First, primary annotators did not have access to model outputs during initial labeling. Second, a senior expert conducted quality control, including targeted review of selected disagreement cases, but final labels were determined through human adjudication according to the refined guidelines rather than by adopting model predictions. Third, evaluation in Section~\ref{sec: results} includes multiple LLMs, so the analysis does not depend only on Claude's behavior. Nevertheless, Claude's scores should be read with this sampling-bias caveat.

\subsection{Annotation Rationales and Prompt Examples}\label{app:annot_rationale}

We present the annotation rationale and prompt templates for \eatmeat to illustrate how labeling guidelines are operationalized in practice.

\paragraph{Annotation Rationale.} 
The annotation rationale presented below reflects the refined guideline used in the final re-annotation stage. An earlier version of the rationale (v1) was used to construct the initial LLM prompts for disagreement sampling (Table~\ref{tab:prompt_eatmeat_initial}). During the re-annotation process, annotators raised recurring corner cases and ambiguities, leading to incremental clarification and expansion of the guideline. The final rationale therefore differs slightly from the initial prompt specification, reflecting standard iterative refinement in human annotation workflows.

The final human annotation guideline for \eatmeat is defined as follows:\\

Label 1 if you see:
\begin{itemizesquish}
    \item Explicit mentions of eating or intentions to eat meat, seafood, or eggs. E.g., ``They had sashimi yesterday''.
    \item Descriptions, recommendations, purchases, or cooking of meat/egg dishes that assume acceptability. E.g., ``The main dish is steak.'', ``Do you like hot pot?''
    \item A negative statement that still implies meat-eating. e.g., ``I don't eat raw fish'' which implies fish is okay when cooked, ''I don't like meat'' which is not absolute rejection.
    \item General references to meat/egg foods as acceptable or desirable. E.g., ``How much is the chicken?'', ``Hot dogs are delicious.''
    \item Special case: mentions of going fishing, e.g., ``I seldom go fishing.''
\end{itemizesquish}

Label 0 if:
\begin{itemizesquish}
\item The speaker clearly rejects all animal products. E.g., ``I'm vegetarian'', ``I don't eat eggs.''
\item It's about dairy-based food, e.g., ice-cream/milk, or generic foods commonly available in vegetarian versions without a clear mention of meat, e.g., ``soup'', ``breakfast''
\item There is no reference to meat, seafood, eggs dishes.
\item A phrase without period/question mark that is purely objective, without any words indicating preference (such as best/delicious), e.g., ``tea, ramen'', ``vegetable or raw fish''
\end{itemizesquish}

\paragraph{Prompt Structure.}
Based on this rationale, we construct prompt templates for different stages of the pipeline. Table~\ref{tab:prompt_eatmeat_initial}, \ref{tab:prompt_eatmeat_norationale} and \ref{tab:prompt_eatmeat} show the prompts used for \eatmeat. In borderline cases, positive labels are preferred to avoid overlooking potential meat-related cues. Other attribute prompts follow a similar design principle with label-specific rationales and precision-recall trade-offs. The finalized prompts released with the data include the label-specific operational requirements used in this paper. They standardize in-context demonstrations using canonical chat-message roles and are used for the public-release evaluation; product-facing versions may continue to evolve.\footnote{For two labels, the released JSON files standardize the chat-role formatting of the in-context demonstration relative to the historical experimental prompts, without changing the demonstration content or system instruction. On the public subset, this change affected GPT-4.1 positive-class F1 by at most 0.01 across the two labels.}

The prompts were developed through manual refinement on a held-out development set,\footnote{LLM tools sometimes assisted prompt drafting and iteration, but final prompt changes were selected manually.} with the goal of producing interpretable and operationally reliable instructions that could be reused across languages. This process provides reliable initial versions of the rationale-derived prompts, which can be further refined through automated prompt-improvement methods as additional validation data and deployment feedback become available.

The prompts generally follow a chat-style format consisting of a system instruction and one or more chat turns. The system message specifies the task definition and output format, while the user message provides the input text to be classified. For evaluation prompts (Tables~\ref{tab:prompt_eatmeat_norationale} and \ref{tab:prompt_eatmeat}), most prompts include a single manually written in-context demonstration before the target query to improve output-format control.\footnote{This demonstration is omitted for simpler prompts where the desired output format can be specified sufficiently in the system and/or user instructions.} In contrast, the sampling prompt (Table~\ref{tab:prompt_eatmeat_initial}) is applied in a zero-shot setting and does not include an in-context example.

\begin{table}[]
\centering
\small
\noindent\fbox{
\begin{minipage}{\dimexpr\linewidth-2\fboxsep-2\fboxrule}
\tt
You are a multilingual content reviewer tasked with identifying whether a sentence refers to or implies the consumption of animal-based foods. The goal is to exclude sentences that are not suitable for vegans — people who avoid all animal products, including meat, poultry, fish, eggs, and dairy. Mislabeling will cause serious risks, please carefully evaluate each text (sentence, short text, or dialogue) according to the following criteria.\\
\\
Please assign a label of 1 (non-vegan) if any of the following are true:\\
1. The speaker explicitly eats or expresses intent to eat meat, seafood, or eggs. E.g., “I had sashimi yesterday”.\\
2. The sentence describes, recommends, buys, or cooks meat/egg dishes in a way that assumes the speaker or user is okay with meat. E.g., “The main dish is steak.”, “Do you like hot pots?”, “Let's grill beef.”\\
3. The sentence includes a negative statement that still implies meat-eating, e.g., “I don't eat raw fish” which implies fish is okay when cooked. “ I don’t like boiled eggs” implies eggs are consumed otherwise.\\
4. The sentence contains general or descriptive references to meat or egg-based food that assumes it's acceptable or desirable to the audience. E.g., “How much is the chicken?”, “Hot dogs are delicious.”\\
\\
Please assign a label of 0 only if:\\
- The speaker clearly reject animal products consumption. E.g., “I'm vegetarian”, “I don't eat eggs.”\\
- There is no reference to meat, seafood, eggs, dairy, or animal-based dishes.\\
\\
If a sentence mentions food, but the context is unclear, you should still label 1 to err on the side of protecting vegetarian users from exposure. Please provide a rationale and a score using the following format:\\
RATIONALE:\\
SCORE:
\end{minipage}
}
\caption{Initial prompt (system instruction) for \eatmeat used for sampling in Section~\ref{sec:sampling}, with Claude 3.7 Sonnet (20250219) at temperature 0.}
\label{tab:prompt_eatmeat_initial}
\end{table}

\begin{table}[]
\centering
\small
\noindent\fbox{
\begin{minipage}{\dimexpr\linewidth-2\fboxsep-2\fboxrule}
\tt
You are a multilingual content reviewer. Your task is to determine whether a sentence implies the speaker is okay with consuming meat-related foods (meat, poultry, seafood, eggs), do not consider dairy-based food. Assign 1 if the speaker is okay with consuming meat-related foods, otherwise assign 0.\\
\\
If unsure, assign SCORE = 1.\\
\\
Output format:\\
RATIONALE: \\
SCORE: 0 or 1
\end{minipage}
}
\caption{Prompt (system instruction) for \eatmeat after refinement on re-annotated data without rationales, used in Section~\ref{sec: results}.}
\label{tab:prompt_eatmeat_norationale}
\end{table}

\begin{table}[]
\centering
\small
\noindent\fbox{
\begin{minipage}{\dimexpr\linewidth-2\fboxsep-2\fboxrule}
\tt
You are a multilingual content reviewer. Your task is to determine whether a sentence implies the speaker is okay with consuming meat-related foods, defined as follows:\\
\\
Meat-related foods include:\\
- Meat, poultry, seafood, or eggs.\\
- Dishes typically made with these ingredients by default or named after them, e.g., "hamburger," "sushi," "ramen," "hot dog."\\
\\
Do NOT treat the following as meat-related:\\
- Foods that are plant-based, e.g., vegetables, tofu, fruit.\\
- Foods that are dairy-based, e.g., cheese, cake, milk, ice cream.\\
- Generic foods or dishes commonly available in vegetarian versions without a clear mention of meat, e.g., soup, breakfast, sandwich, kimbap.\\
\\
Assign SCORE = 1 if ANY of these conditions apply:\\
\\
- The speaker explicitly eats, intends to eat, cooks, buys, recommends, or positively evaluates meat-related foods. Examples: "I had sashimi yesterday," "Let's grill beef," "Hot dogs are delicious.".\\
- Neutral or descriptive sentences about meat-related food or places, implying potential speaker involvement or intent to consume, e.g., "5 minutes walk from the BBQ place", "That is a hotdog shop." \\
- The speaker or someone else consumes meat-related foods (even indirectly). Examples: "My friend had pork," "Cats eat fish."\\
- The sentence expresses partial rejection that implies general acceptance. Examples: "I don’t eat raw fish" (implying cooked fish is acceptable), "I don't like pork" (suggesting the speaker can still consume pork).\\
- The sentence explicitly mentions activities that imply obtaining meat-related food, e.g., "I seldom go fishing," "She likes to go fishing."\\
- Questions or offers explicitly including meat as an acceptable option, e.g., "Would you like some chicken?"\\
- The sentence is a short phrase that clearly implies speaker preference or positive evaluation, e.g., "delicious fish," "best sushi shop," "favorite hotdog."\\
\\
Assign SCORE = 0 ONLY if:\\
- The speaker explicitly rejects consuming meat, seafood, eggs, or animal products. Examples: "I'm vegetarian", "I don't eat meat", "No eggs for me.", "I never go fishing."\\
- The sentence is a short PHRASE that does NOT clearly imply meat consumption or preference. Examples: "meat shop", "pork or lamb", "cold hotdog and tea."\\
\\
If unsure, assign SCORE = 1.\\
\\
Output format:\\
RATIONALE:\\
SCORE: 0 or 1
\end{minipage}
}
\caption{Prompt (system instruction) for \eatmeat after refinement on re-annotated data, used in Section~\ref{sec: results}.}
\label{tab:prompt_eatmeat}
\end{table}

\subsection{Comparison with Original Labels} \label{app:agreement}

Table~\ref{kappa} shows Cohen's $\kappa$ between the original noisy labels from Section~\ref{sec:original_sample} and the re-annotated labels from Section~\ref{sec:reannotation}, broken down by language and label.

\begin{table*}[!t]
\small
\centering
\begin{tabular}{p{0.09\linewidth}p{0.07\linewidth}p{0.07\linewidth}p{0.07\linewidth}p{0.07\linewidth}p{0.07\linewidth}p{0.07\linewidth}p{0.09\linewidth}p{0.07\linewidth}p{0.07\linewidth}}\toprule
 language & \male & \female & \child & \adult & \grandparent & \parent & \eatmeat & \vegetarian & \serious \\\midrule
Japanese & 0.91 & 1.00 & 0.12 & 0.27 & 0.91 & 1.00 & 0.62 & 0.67 & 0.80 \\
Portuguese & 0.33 & 0.09 & 0.02 & 0.38 & 0.74 & 0.60 & 0.80 & 0.83 & 0.00 \\
English & 0.03 & 0.04 & 0.02 & 0.28 & 0.58 & 0.02 & 0.86 & 0.47 & 0.17 \\
Spanish & 0.20 & 0.08 & 0.52 & 0.59 & 0.83 & 0.38 & 0.75 & 0.58 & 0.10 \\
German & 0.14 & 0.13 & 0.10 & 0.18 & 1.00 & 0.34 & 0.92 & 0.73 & 0.24 \\
French & 0.75 & 0.84 & 0.01 & 0.35 & 0.73 & 0.28 & 0.90 & 0.45 & 0.71 \\
Italian & 1.00 & 0.92 & 0.02 & 0.68 & 0.67 & 0.82 & 0.80 & 0.35 & 0.81 \\
Korean & 0.31 & 0.13 & 0.51 & 0.29 & 0.80 & 0.85 & 0.82 & 0.69 & 0.46 \\
Russian & 0.20 & 0.35 & 0.07 & 0.47 & 0.66 & 0.82 & 0.92 & 0.30 & 0.80 \\
Turkish & 0.23 & 0.15 & 0.10 & 0.35 & 0.92 & 0.53 & 0.59 & 0.60 & 0.15 \\
Chinese & 0.07 & 0.09 & 0.13 & 0.12 & 1.00 & 0.84 & 0.77 & 0.53 & 0.18 \\\midrule 
Total & 0.23 & 0.17 & 0.12 & 0.36 & 0.80 & 0.45 & 0.78 & 0.56 & 0.33 \\\bottomrule
\end{tabular}
\caption{\label{kappa} Cohen's $\kappa$ per language and label.}
\end{table*}

\subsection{Additional Evaluation Metrics on the Re-annotated Test Set}

Table~\ref{tab:macro_f1_reannotated} reports macro-F1 on the re-annotated test set as a complementary metric to the positive-class F1 results in Table~\ref{results}. The relative trends are consistent with the main results: Claude 3.7 Sonnet performs best on five of the nine labels, while GPT-4.1 remains a close second.

\begin{table*}[!t]
\small
\centering
\begin{tabular}{p{0.15\linewidth}|p{0.06\linewidth}p{0.06\linewidth}|p{0.06\linewidth}p{0.06\linewidth}p{0.06\linewidth}|p{0.06\linewidth}|p{0.09\linewidth}p{0.06\linewidth}|p{0.06\linewidth}}\toprule
\multirow{2}{*}{} & \multicolumn{2}{c|}{gender} & \multicolumn{3}{c|}{age group} &   & \multicolumn{2}{c|}{diet} &  \\
 & \male & \multicolumn{1}{l|}{\female} & \child & \multicolumn{1}{l}{\adult} & \multicolumn{1}{l|}{\grandparent} & \parent & \eatmeat & \multicolumn{1}{l|}{\vegetarian} & \serious \\\midrule
DeepSeek V3 & 86.2 & 82.3 & 66.9 & 84.1 & 84.5 & 88.1 & 89.9 & 78.2 & 74.1 \\
Gemini 2.5 Flash & 90.9 & 88.3 & 66.5 & 81.6 & 81.2 & 83.9 & 90.1 & \textbf{92.8} & 69.5 \\
GPT 4.1 & \textbf{91.9} & 94.1 & 69.0 & 86.5 & 91.5 & \textbf{92.9} & \textbf{91.7} & 85.7 & 75.3 \\
Claude 3.7 Sonnet & 91.9 & \textbf{96.5} & \textbf{70.0} & \textbf{86.8} & \textbf{94.5} & 92.4 & 90.3 & 90.9 & \textbf{78.6}\\
\bottomrule
\end{tabular}
\caption{\label{tab:macro_f1_reannotated} Macro-F1 (\%) on the re-annotated test set. Bold indicates the highest score for each label before rounding.}
\end{table*}

\subsection{Inter-Model Agreement}\label{app:fleiss}

\begin{table*}[!t]
\small
\centering
\begin{tabular}{p{0.09\linewidth}p{0.07\linewidth}p{0.07\linewidth}p{0.07\linewidth}p{0.07\linewidth}p{0.07\linewidth}p{0.07\linewidth}p{0.09\linewidth}p{0.07\linewidth}p{0.07\linewidth}}\toprule
 language & \male & \female & \child & \adult & \grandparent & \parent & \eatmeat & \vegetarian & \serious \\\midrule
Japanese & 0.90 & 0.93 & 0.56 & 0.71 & 0.78 & 0.95 & 0.91 & 0.72 & 0.79 \\
Portuguese & 0.81 & 0.68 & 0.64 & 0.86 & 0.90 & 0.79 & 0.93 & 0.59 & 0.78 \\
English & 0.39 & 0.79 & 0.62 & 0.85 & 0.46 & 0.79 & 0.83 & 0.47 & 0.72 \\
Spanish & 0.73 & 0.66 & 0.59 & 0.84 & 0.76 & 0.88 & 0.89 & 0.82 & 0.62 \\
German & 0.65 & 0.66 & 0.69 & 0.72 & 0.79 & 0.75 & 0.95 & 0.75 & 0.72 \\
French & 0.80 & 0.67 & 0.51 & 0.78 & 0.61 & 0.75 & 0.85 & 0.75 & 0.74 \\
Italian & 0.85 & 0.85 & 0.68 & 0.73 & 0.85 & 0.77 & 0.87 & 0.71 & 0.75 \\
Korean & 0.63 & 0.64 & 0.48 & 0.70 & 0.80 & 0.78 & 0.91 & 0.71 & 0.67 \\
Russian & 0.82 & 0.54 & 0.54 & 0.80 & 0.57 & 0.82 & 0.98 & 0.82 & 0.78 \\
Turkish & 0.75 & 0.90 & 0.63 & 0.70 & 0.59 & 0.71 & 0.90 & 0.70 & 0.66 \\
Chinese & 0.57 & 0.62 & 0.63 & 0.87 & 0.77 & 0.63 & 0.91 & 0.75 & 0.70 \\\midrule 
Total & 0.74 & 0.71 & 0.60 & 0.79 & 0.73 & 0.77 & 0.90 & 0.71 & 0.72 \\\bottomrule
\end{tabular}
\caption{\label{fleiss} Inter-model Fleiss' $\kappa$ across the four benchmarked LLMs, per language and label.}
\end{table*}

Table~\ref{fleiss} reports inter-model Fleiss' $\kappa$ across the four benchmarked LLMs on the re-annotated test set, measuring how consistently the models apply the refined evaluation prompts to the same instances.

Agreement is highest on \eatmeat ($\kappa = 0.90$) and lowest on \child ($0.60$). Manual inspection suggests that many disagreements reflect conflicts between the refined rationales and surface-based readings, including over-labeling of indirect family references for \parent and \grandparent, partner or address terms for \male and \female, and mild negative or complaint-like phrasings for \serious. In the inspected disagreement cases, Claude and GPT more often follow the refined exclusions, whereas Gemini and DeepSeek are more prone to over-labeling based on surface cues, illustrating the lexical-overreliance limitation discussed in Section~\ref{sec:discussion}. A different pattern appears for \vegetarian, where GPT-4.1 sometimes treats questions such as ``Is this pizza vegetarian?'' as speaker-neutral despite the operational rationale treating them as evidence of vegetarian preference.

\subsection{Results on Public Release Subset}
For reproducibility on the released public subset, we report GPT-4.1 results using the released final prompts. We choose GPT-4.1 as a strong reference model that was not used in disagreement-focused sampling, and because Claude 3.7 Sonnet, used in our main experiments, is no longer available.\footnote{Retired by Anthropic on February 19, 2026.} These results are intended as a reference point and are not directly comparable to the main benchmark results due to differences in sampling distribution and language coverage.

Table~\ref{stats_public} reports the positive-class support for each label in the public release subset. Table~\ref{results_public_f1} reports positive-class F1, matching the primary metric used in Table~\ref{results}. Table~\ref{results_public_macro} additionally reports macro-F1 as a secondary summary over both classes. In both tables, EN, ES, IT, KO, and ZH correspond to English, Spanish, Italian, Korean, and Chinese, respectively; the \textit{all} column reports F1 computed on the pooled five-language slice for each label, using a single confusion matrix over all released instances for that label. The \textit{avg} row reports the arithmetic mean of the nine per-label scores in each column.

The relatively narrow per-language spread under the refined prompts contrasts with the wide cross-lingual variability observed in the original annotations (Table~\ref{kappa}), suggesting that rationale-based refinement improves cross-lingual consistency.

\begin{table}[!t]
\small
\centering
\begin{tabular}{p{0.18\linewidth}p{0.18\linewidth}p{0.18\linewidth}p{0.18\linewidth}}\toprule
 & positives & negatives & total \\\midrule
\male       & 96  & 304 & 400 \\
\female     & 69  & 331 & 400 \\
\child      & 116 & 284 & 400 \\
\adult      & 199 & 201 & 400 \\
\grandparent & 93 & 307 & 400 \\
\parent     & 171 & 229 & 400 \\
\eatmeat    & 194 & 206 & 400 \\
\vegetarian & 91  & 309 & 400 \\
\serious    & 200 & 200 & 400 \\\midrule
total       & 1{,}229 & 2{,}371 & 3{,}600 \\\bottomrule
\end{tabular}
\caption{Per-label positive- and negative-class support in the public release subset.}
\label{stats_public}
\end{table}

\begin{table}[!t]
\small
\centering
\begin{tabular}{p{0.18\linewidth}p{0.07\linewidth}p{0.07\linewidth}p{0.07\linewidth}p{0.07\linewidth}p{0.07\linewidth}|p{0.06\linewidth}}\toprule
  & EN & ES & IT & KO & ZH & all \\\midrule
\male        & 100.0 & 80.8 & 88.9 & 88.9 & 90.9 & 87.8 \\
\female      &  87.5 & 90.9 & 98.0 & 80.0 & 92.3 & 91.0 \\
\child       &  82.6 & 80.0 & 80.0 & 78.9 & 81.8 & 80.7 \\
\adult       &  90.4 & 88.7 & 71.7 & 85.7 & 87.1 & 86.1 \\
\grandparent & 100.0 & 91.7 & 89.4 & 93.6 & 90.0 & 93.2 \\
\parent      &  98.8 &100.0 & 93.8 & 95.8 & 96.8 & 97.6 \\
\eatmeat     &  96.0 & 93.8 & 93.6 & 95.8 & 92.8 & 94.3 \\
\vegetarian  &  69.2 & 90.3 & 88.4 & 89.7 & 86.7 & 85.5 \\
\serious     &  93.2 & 90.9 & 83.6 & 89.4 & 92.2 & 90.6 \\\midrule
avg          &  90.9 & 89.7 & 87.5 & 88.7 & 90.1 & 89.7 \\\bottomrule
\end{tabular}
\caption{\label{results_public_f1} GPT-4.1 F1 (\%) for the positive class on the public release subset.}
\end{table}

\begin{table}[!t]
\small
\centering
\begin{tabular}{p{0.18\linewidth}p{0.07\linewidth}p{0.07\linewidth}p{0.07\linewidth}p{0.07\linewidth}p{0.07\linewidth}|p{0.06\linewidth}}\toprule
 & EN & ES & IT & KO & ZH & all \\\midrule
\male        & 100.0 & 87.0 & 87.9 & 89.9 & 94.9 & 92.0 \\
\female      &  93.2 & 94.9 & 98.0 & 85.7 & 95.6 & 94.5 \\
\child       &  88.7 & 86.7 & 81.8 & 83.0 & 88.3 & 86.6 \\
\adult       &  90.0 & 89.0 & 69.9 & 86.0 & 88.7 & 86.5 \\
\grandparent & 100.0 & 95.3 & 90.0 & 94.0 & 94.4 & 95.6 \\
\parent      &  99.0 &100.0 & 95.4 & 96.0 & 97.0 & 98.0 \\
\eatmeat     &  96.0 & 94.0 & 94.0 & 96.0 & 93.0 & 94.5 \\
\vegetarian  &  82.3 & 94.3 & 89.8 & 92.7 & 92.2 & 91.0 \\
\serious     &  93.0 & 91.0 & 81.8 & 90.0 & 92.0 & 90.5 \\\midrule
avg          &  93.6 & 92.5 & 87.6 & 90.4 & 92.9 & 92.1 \\\bottomrule
\end{tabular}
\caption{\label{results_public_macro} GPT-4.1 macro-F1 (\%) on the public release subset.}
\end{table}

\end{document}